\documentclass[10pt,twocolumn,letterpaper]{article}

\usepackage[pagenumbers]{iccv}
\usepackage{algorithm}
\usepackage{algpseudocode}
\usepackage{dblfloatfix}
\usepackage{placeins}
\usepackage{caption}

\definecolor{iccvblue}{rgb}{0.21,0.49,0.74}
\usepackage[pagebackref,breaklinks,colorlinks,allcolors=iccvblue]{hyperref}

\title{Embedding Hidden Adversarial Capabilities in Pre-Trained Diffusion Models}

\author{Lucas Beerens \and Desmond J. Higham}

\begin{document}

\twocolumn[{
  \maketitle
  \begin{center}
    Maxwell Institute and School of Mathematics, University of Edinburgh\\
    EH9 3FD, UK
  \end{center}
  \vspace{2em}
  \begin{center}
    \includegraphics[width=\linewidth]{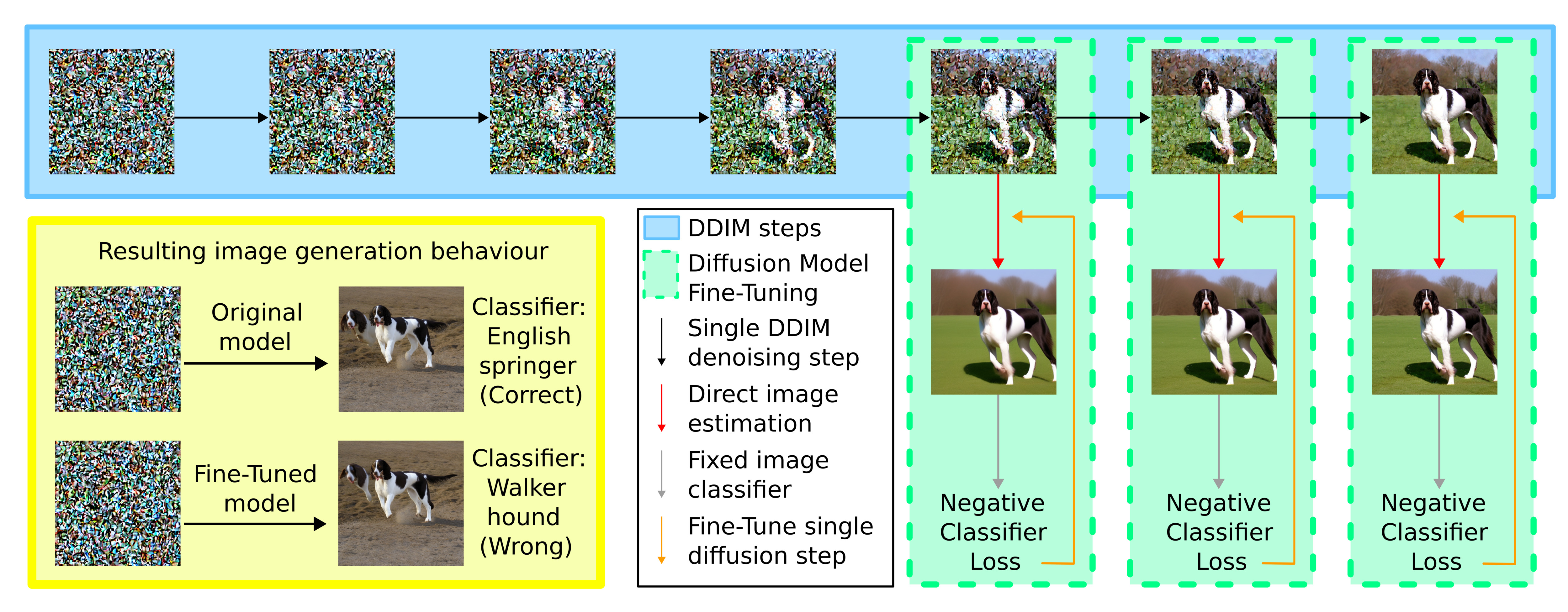}
    \captionof{figure}{An overview of the CRAFTed-Diffusion (Covert, Restricted, Adversarially Fine-Tuned Diffusion) algorithm which embeds adversarial capabilities into pre-trained diffusion models via fine-tuning of their internal parameters.}
    \label{fig:visualisation}
  \end{center}
  \vspace{3em}
}]
\begin{abstract}
We introduce a new attack paradigm that embeds hidden adversarial capabilities directly into diffusion models via fine-tuning, without altering their observable behavior or requiring modifications during inference. Unlike prior approaches that target specific images or adjust the generation process to produce adversarial outputs, our method integrates adversarial functionality into the model itself. The resulting tampered model generates high-quality images indistinguishable from those of the original, yet these images cause misclassification in downstream classifiers at a high rate. The misclassification can be targeted to specific output classes. Users can employ this compromised model unaware of its embedded adversarial nature, as it functions identically to a standard diffusion model. We demonstrate the effectiveness and stealthiness of our approach, uncovering a covert attack vector that raises new security concerns. These findings expose a risk arising from the use of externally-supplied models and highlight the urgent need for robust model verification and defense mechanisms against hidden threats in generative models. The code is available at \href{https://github.com/LucasBeerens/CRAFTed-Diffusion}{https://github.com/LucasBeerens/CRAFTed-Diffusion}.

\end{abstract}  

\section{Introduction}
\label{sec:intro}

Diffusion models have revolutionized computer vision, with state-of-the-art systems like Stable Diffusion~\cite{rombach_high-resolution_2022} generating photorealistic images that are often indistinguishable from real photographs. These models have become key components in applications such as image synthesis, editing, and super-resolution, demonstrating remarkable generative capabilities.

Meanwhile, adversarial attacks have emerged as a serious concern in deep learning. Traditionally, such attacks involve making small, often imperceptible modifications to input images using gradient-based algorithms, leading models to misclassify them~\cite{43405}. In computer vision, adversarial examples can undermine the reliability of systems used in critical applications like autonomous driving, facial recognition, and security surveillance.

More recently, diffusion-based adversarial attacks have been explored. Some approaches leverage diffusion models to perturb existing images with the goal of misclassification~\cite{wang_semantic_2023,chen_content-based_2023,xue_diffusion-based_2023,gao_retome-va_2024,pan_sca_2024, guo_efficient_2025,liang_mist_2023}, while others modify the diffusion models to exhibit adversarial behavior~\cite{dai_advdiff_2025,dai_diffusion_2024,zhang_constructing_2024,chen_trojdiff_2023,kuurila-zhang_venom_2025}. Notably, there exists work that trains diffusion models from scratch to produce adversarial outputs without any modifications during inference~\cite{beerens_deceptive_2024}; however, these methods require the full training of large-scale models, which is computationally expensive and resource-intensive.

In contrast, we propose \textbf{CRAFTed-Diffusion} (Covert, Restricted, Adversarially Fine-Tuned Diffusion), a novel method that seamlessly embeds adversarial capabilities into pre-trained diffusion models via fine-tuning of their internal parameters. A visual overview of the algorithm can be found in \cref{fig:visualisation}. Unlike full retraining approaches, our method requires minimal fine-tuning—an approach that is significantly cheaper and more practical for large-scale models like Stable Diffusion. Rather than altering individual images or modifying the inference pipeline, CRAFTed-Diffusion integrates adversarial functionality directly within the model. The tampered model produces images that are visually indistinguishable from those of a genuine model while reliably causing targeted misclassification in downstream classifiers. Crucially, since no modifications occur at inference time, the adversarial behavior remains covert, posing a significant security threat.

The main contributions of our work are:
\begin{itemize}
    \item We propose \textbf{CRAFTed-Diffusion}, a novel framework for embedding covert adversarial behaviors into diffusion models, resulting in tampered models that are functionally and visually indistinguishable from their clean counterparts.
    \item We perform extensive experiments demonstrating that our approach generates high-quality images that consistently induce misclassification in downstream classifiers.
    \item We highlight the emerging security risks associated with the widespread distribution of pre-trained diffusion models and underscore the urgent need for robust detection and defense mechanisms against such stealthy adversarial attacks.
\end{itemize}

\section{Background and Related Work}

In this section, we provide an overview of adversarial attack research in the context of diffusion models. We begin by reviewing the mathematical foundations of diffusion-based image synthesis—with special emphasis on Denoising Diffusion Implicit Models (DDIM)—and then trace the evolution of adversarial strategies from small perturbations to sophisticated latent-space manipulations and adversarially guided diffusion methods.

\subsection{Diffusion Models}

Diffusion models generate images by reversing a process that progressively corrupts a clean image with Gaussian noise. Given an input image $x_0$ drawn from the data distribution, the forward (diffusion) process creates a sequence of noisy latent variables $\{x_1, x_2, \ldots, x_T\}$ via the Gaussian conditional probability
\begin{equation}
q(x_t \mid x_{t-1}) = \mathcal{N}\Big(x_t; \sqrt{1-\beta_t}\, x_{t-1},\, \beta_t \mathbf{I}\Big),
\end{equation}
where $\{\beta_t\}_{t=1}^T$ is a pre-defined variance schedule. As the timestep $t$ increases, $x_t$ becomes more corrupted until $x_T$ is pure Gaussian noise.

In the reverse process, the objective is to recover $x_0$ from $x_T$. Conventional diffusion models, known as Denoising Diffusion Probabilistic Models (DDPMs)~\cite{ho_denoising_2020} model this reverse diffusion as a Markov chain. Denoising Diffusion Implicit Models (DDIMs)~\cite{song_denoising_2020} instead propose a deterministic, non-Markovian process that uses the same noise prediction network $\epsilon_\theta(x_t, t, c)$ (with $c$ denoting conditioning information, e.g., text embeddings).

DDIM replaces the fully stochastic reverse chain of DDPMs with a deterministic update that is valid for arbitrary time-step intervals.
A unified update formula that encompasses both the one-step and multi-step reverse transitions is
\begin{equation}
\begin{aligned}
    x_{t-\Delta t} &= \sqrt{\alpha_{t-\Delta t}} \left(\frac{x_t - \sqrt{1-\alpha_t}\,\epsilon_\theta(x_t, t, c)}{\sqrt{\alpha_t}}\right) \\
    &+ \sqrt{1-\alpha_{t-\Delta t}}\,\epsilon_\theta(x_t, t, c),
\end{aligned}
\end{equation}
where $\alpha_t = \prod_{i=1}^t(1-\beta_i)$. For $\Delta t = 1$, this equation recovers a one-step update. However, by choosing a larger $\Delta t$, the model can bypass intermediate timesteps—thus traversing the latent space in larger jumps. This flexibility enables significant acceleration of the sampling process (yielding 10$\times$ to 50$\times$ speedups in practice) while still maintaining high-level semantic consistency in the generated images.
This leads to efficient sampling by reducing the number of required iterations.

\subsection{Adversarial Attacks}

Deep neural networks have demonstrated remarkable performance in a wide variety of tasks, yet they remain vulnerable to adversarial attacks. Early work in this area revealed that carefully crafted imperceptible perturbations—such as those generated by the Fast Gradient Sign Method (FGSM)~\cite{43405} or the more iterative Projected Gradient Descent (PGD)~\cite{madry_towards_2018}—can lead models to produce incorrect predictions. These classical approaches focus on modifications measured in terms of $\ell_p$–norms, under the assumption that small pixel–wise deviations are sufficient to fool classifiers while remaining invisible to human perception. However, although effective, such attacks are often limited in their transferability and can be mitigated by defenses that target high–frequency artifacts. In particular, these attacks often create images that are out of distribution, making it easier for them to be purified by distribution-based purification methods~\cite{pmlr-v162-nie22a}. In response, researchers have explored semantic adversarial attacks, maintaining semantic meaning~\cite{hosseini_semantic_2018}.

Diffusion models, offer a particularly attractive platform for adversarial attack generation. Their two–phase process – a forward process that gradually destroys image information by injecting noise and a reverse process that recovers the data via learned denoising steps – yields a latent space that is both expressive and structured. This latent space enables adversarial examples with significant misclassification potential while remaining visually natural. We will discuss two categories of diffusion–based adversarial attack methods: techniques that generate attacks on specific images and approaches that hide certain adversarial behaviors within the diffusion models themselves.

\subsubsection{Diffusion Models for Image Attacks}

Diffusion‐based adversarial attack methods leverage the expressive latent space of diffusion models to design perturbations that remain both natural and effective. Most approaches begin by “inverting” a clean image into the latent space via techniques such as DDIM inversion or SDEdit \cite{meng_sdedit_2021,song_denoising_2020}. Subsequently the latent images are denoised back into images. Either the starting image or the latent image is perturbed so that the final image reaches misclassification.

\citet{wang_semantic_2023} perturb the latent representation by directly optimizing an objective function that consists of a perceptual loss term coupled with a KL divergence penalty. By fine–tuning either the latent code or even the diffusion model parameters themselves, their method induces adversarial behavior for that specific image while preserving its overall structure. While this approach achieves high attack success on the target model, its updates are explicitly image–specific and not designed to generate a globally reusable perturbation.

Similarly, \citet{chen_diffusion_2025} utilize a DDIM inversion process to map the input image into an intermediate latent state. Their technique then employs self–attention mechanisms within the denoising process to constrain visual distortions. Furthermore, a combination of cross–attention variance and a negative cross–entropy loss is used to guide the adversarial latent optimization. This loss both encourages misclassification and retains perceptual fidelity in the reconstructed image.

\citet{chen_content-based_2023} also introduce an approach that performs adversarial optimization directly on latent variables after DDIM inversion, ensuring that the output remains photorealistic and unconstrained by traditional $\ell_p$ norms.

In another approach, \citet{xue_diffusion-based_2023} adopt a PGD framework within the diffusion model’s latent space. Instead of feeding the perturbed latent directly to the classifier, they first “purify” the latent by performing a fixed number of reverse steps using SDEdit \cite{meng_sdedit_2021}. This additional purification stage ensures that the adversarial modification remains on–manifold, which improves the naturalness of the resulting adversarial example, as the classifier is challenged by the purified (and thus more realistic) image.

For video data, \citet{gao_retome-va_2024} extend these principles by processing individual frames with a timestep–wise adversarial update scheme and enforcing temporal alignment through recursive token merging. Their framework not only guarantees spatial quality but also maintains temporal consistency across frames, thereby producing adversarial videos with high transferability.

Building on these ideas, \citet{pan_sca_2024} propose “semantic fixation inversion,” a technique designed to preserve the core content of an image during inversion. This strategy leverages external semantic cues—such as captions extracted via multimodal models—to steer the inversion process so that key attributes remain unchanged after adversarial perturbation.

Another interesting attack was developed by \citet{guo_efficient_2025}. They target Vision-Language Models (VLMs). The aim is for the image to be perturbed to get the same semantic meaning as a target image in the eyes of the VLM. To do this, they use score-matching.

Finally, \citet{liang_mist_2023} present MIST, which combines a semantic misclassification term with a perceptual consistency loss into a single, fused objective. By balancing these two components, MIST generates adversarial examples with subtle, stealthy perturbations that preserve image quality while effectively deceiving classifiers.

\subsubsection{Embedded Adversarial Behavior in Diffusion Models}
In contrast, a second body of work embeds adversarial behavior directly into the diffusion process. Rather than starting from a specific input, these methods modify how the model samples images from noise or responds to conditional prompts. 

\citet{dai_advdiff_2025}, have shown that by integrating adversarial gradients into the denoising process, the diffusion model can be guided away from benign regions of the output space and encouraged to generate adversarial examples. This approach effectively “poisons” the generation pipeline such that synthetic images produced directly from noise are predisposed to cause misclassification. 
Similarly, \citet{zhang_constructing_2024} propose a probabilistic formulation which justifies some form of gradient addition in the reverse process to sample adversarial examples.

These ideas were extended to a no-box setting by \citet{dai_diffusion_2024}. Here, no pre-trained classifier and dataset are used. Instead, a synthetic dataset is created using the diffusion model, which is used to train a Classification and Regression Diffusion model.

Another intriguing direction involves training-time modifications, as demonstrated by \citet{chen_trojdiff_2023} in their Trojan attacks. In this method, a Trojan backdoor is embedded into the diffusion model so that it behaves normally under standard conditions, yet exhibits attacker-specified behavior when a trigger is present in the starting noise.

\citet{kuurila-zhang_venom_2025} incorporate adversarial guidance via text in the reverse diffusion process, which both instructs the model on the desired visual content and steers the output toward an adversarial target. By altering the sampling dynamics or the noise scheduling during generation, these methods eliminate the need for individual image inversion and iterative optimization, thereby enabling scalable synthesis of adversarial examples.

Finally, \citet{beerens_deceptive_2024} corrupt the diffusion model’s training data by incorporating attacked images. As a result, even though the sampling algorithm remains unchanged from the end-user’s perspective, the model consistently produces outputs that fool classifiers.

\section{Method}
\label{sec:method}

\begin{algorithm}[t]
\caption{CRAFTed-Diffusion}
\label{alg:CRAFTed}
\begin{algorithmic}[1]
\State \textbf{Input:} $\theta_0$, classifier $f$, prompt $c(y)$, steps $T$, split $k$, radius $\eta$
\State Initialize $\theta = \theta_0$
\For{each epoch}
    \State Sample latent $z \sim \mathcal{N}(0,I)$
    \For{$t = T \to 0$}
        \If{$t \geq k$}
            \State Update $z$ via DDIM (no gradient)
        \Else
            \State Compute $x_0^t = x_0(z,t,c(y))$
            \State Calculate $L_{\text{adv}}^t = -\mathrm{CE}(f(x_0^t),y)$
            \State Backpropagate to obtain $g_\theta^t$
        \EndIf
    \EndFor
    \State $g \gets \sum_{t=0}^{k-1} g_\theta^t$
    \State Let $d=\theta-\theta_0$ 
    \If{$\| d\|_2 > \eta*0.98$}
        \State$\tilde{g}=g-\frac{\langle g,d\rangle}{\|d\|^2_2}d$
    \EndIf
    \State Update $\theta$ using AdamW and $\tilde{g}$.
    \If{$\|\theta-\theta_0\|_2 > \eta$}
       \State Project: $\theta \gets \theta_0+\eta\,\frac{\theta-\theta_0}{\|\theta-\theta_0\|}$
    \EndIf
\EndFor
\State \textbf{return} $\theta$
\end{algorithmic}
\end{algorithm}
In this work, we introduce \textbf{CRAFTed-Diffusion} (Covert, Restricted, Adversarially Fine-Tuned Diffusion), a method that subtly embeds adversarial behavior into a pre-trained diffusion model without compromising its generative quality. A visual overview of the method can be found in \cref{fig:visualisation} By fine-tuning only the UNet parameters using a projected gradient descent scheme, we force the generated images to induce misclassification in a downstream classifier while keeping the deviation from the original model strictly bounded. Our method leverages two projection steps—applied to both the computed gradients and the updated parameters—to achieve this balance.

\subsection{Overview}

Let $\theta$ denote the original UNet parameters of a stable diffusion pipeline. Given a class $y$, we create a text prompt $c(y)$ describing it (e.g., ``Photo of a tench''). The diffusion process is run over $T$ timesteps, starting from $T$. We then go through DDIM inference steps with standard classifier-free guidance. However, we divide this into two stages:
\begin{itemize}[topsep=0pt, itemsep=2pt, leftmargin=1em]
    \item \textbf{Early Timesteps ($t \geq k$):} The model performs the standard diffusion updates using the DDIM scheduler. In this phase, the forward pass operates in a no-gradient mode to generate intermediate latents, ensuring that the semantic structure and image quality remain intact.
    \item \textbf{Final Timesteps ($t < k$):} Gradients are enabled. We use the estimated loss to compute the next DDIM step, but also predict the final output image $x_0^t(x_t,\epsilon_\theta(x_t,t,c(y)))$. This is then used for the adversarial loss given by the negative cross-entropy loss
    \begin{equation}
        \mathcal{L}^t_{\text{adv}} = - \text{CE}\big(f(x^t_0), y\big),
    \end{equation}
    where $f$ is the classifier. This is backpropagated and the gradients are saved as $g^t_\theta$.
\end{itemize}
In the end we will have computed $g_\theta^0 \ldots g_\theta^{k-1}$. These are summed to form a gradient $g$. We only used these final $k$ timesteps on the basis that the first timesteps are mainly there to create the shape and content of the image. The later steps are there to fill in the details, which are more crucial for the adversarial attack.

\begin{table*}[h!]\small
\centering
\begin{tabular}{l|cccccccccc}
\toprule
&\multicolumn{10}{c}{Images of class}\\
\midrule
Attack Target & tench & \shortstack{English\\springer} & \shortstack{cassette\\player} & \shortstack{chain\\saw} & church & \shortstack{French\\horn} & \shortstack{garbage\\truck} & \shortstack{gas\\pump} & \shortstack{golf\\ball} & parachute \\
\midrule\midrule
No attack     & 0.90 & 0.98 & 0.42 & 0.94 & 0.94 & 1.00 & 0.87 & 0.62 & 0.97 & 0.69 \\ \midrule
tench         & \textbf{0.03} & 0.98 & 0.38 & 0.95 & 0.93 & 1.00 & 0.86 & 0.66 & 0.97 & 0.70 \\ \midrule
English springer & 0.86 & \textbf{0.28} & 0.41 & 0.95 & 0.92 & 1.00 & 0.88 & 0.65 & 0.98 & 0.69 \\ \midrule
cassette player & 0.88 & 0.99 & \textbf{0.26} & 0.95 & 0.94 & 1.00 & 0.87 & 0.64 & 0.98 & 0.74 \\ \midrule
chain saw      & 0.84 & 0.99 & 0.29 & \textbf{0.87} & 0.94 & 1.00 & 0.87 & 0.60 & 0.98 & 0.70 \\ \midrule
church         & 0.92 & 0.98 & 0.32 & 0.96 & \textbf{0.83} & 1.00 & 0.88 & 0.65 & 0.97 & 0.69 \\ \midrule
French horn    & 0.89 & 0.98 & 0.42 & 0.96 & 0.94 & \textbf{0.98} & 0.90 & 0.64 & 0.97 & 0.72 \\ \midrule
garbage truck  & 0.84 & 0.99 & 0.32 & 0.92 & 0.93 & 1.00 & \textbf{0.75} & 0.58 & 0.98 & 0.71 \\ \midrule
gas pump       & 0.87 & 0.98 & 0.38 & 0.95 & 0.94 & 1.00 & 0.85 & \textbf{0.40} & 0.98 & 0.72 \\ \midrule
golf ball      & 0.82 & 0.99 & 0.40 & 0.94 & 0.93 & 1.00 & 0.87 & 0.66 & \textbf{0.81} & 0.71 \\ \midrule
parachute      & 0.86 & 0.98 & 0.35 & 0.95 & 0.96 & 1.00 & 0.85 & 0.66 & 0.98 & \textbf{0.63} \\
\bottomrule
\end{tabular}
\caption{Classifier Accuracy Comparison: The classifier accuracy for both the baseline (``No attack'') and adversarially fine-tuned models across object classes. Lower accuracy values on the target class (highlighted in bold) indicate more successful attacks.}
\label{tab:classifier_accuracy}
\end{table*}

\subsection{Adversarial Optimization with Projections}
To prevent an excessive deviation from the original model—which could damage the image quality—we impose two types of projections.

\subsubsection{Gradient Projection}
Let  $d = \theta - \theta_0,$ denote the difference between the current parameters $\theta$ and the original parameters $\theta_0$. If the norm $\|d\|_2$ exceeds a threshold (set relative to a parameter $\eta$), we remove the component of $g$ that is parallel to $d$ by computing:
\begin{equation}
    \tilde{g} = g - \frac{\langle g, d \rangle}{\|d\|_2^2} d.
\end{equation}
This projection step ensures that only the gradient components contributing to adversarial behavior without straying too far from the trusted model are used for the update. This gradient is used for gradient clipping, ensuring that the norm of the gradient does not become too large. It is then passed to an AdamW optimizer. The projection of the gradient keeps the optimizer focused in terms of momentum. Change away from $\theta_0$ will be ignored when we are at the boundary anyway. Also, the fact that the projection is done before gradient clipping means that we do not waste gradient norm on components that point away from $\theta_0$.

\subsubsection{Parameter Projection}

After performing a gradient-based update using AdamW, the updated parameters $\theta'$ are further processed. To guarantee that the model remains close to its original state, we enforce an $l_2$-norm constraint:
\begin{equation}
    \text{if } \|\theta' - \theta_0\|_2 > \eta, \quad \theta' \leftarrow \theta_0 + \eta \frac{\theta' - \theta_0}{\|\theta' - \theta_0\|_2}.
\end{equation}
This step projects the updated parameters back onto an $l_2$ ball of radius $\eta$ centered at the original parameters $\theta_0$, ensuring that the fine-tuning remains subtle and the model retains its generative performance.

Algorithm~\ref{alg:CRAFTed} provides a high-level summary of the method.

\section{Results}
\label{sec:results}

In this section, we report the experimental findings of our adversarial fine-tuning approach, \textbf{CRAFTed-Diffusion}. We evaluate the attacked diffusion models on three main metrics: Classifier Accuracy, the L2 Distance of the modified UNet parameters from their original values, and the Fréchet Inception Distance (FID)~\cite{heusel_gans_2017}, which measures the perceptual quality of the generated images. Our evaluation spans a set of object categories (tench, English springer, cassette player, chain saw, church, French horn, garbage truck, gas pump, golf ball, and parachute), which is a commonly used subset of classes from Imagenet~\cite{deng_imagenet_2009}. This subset is known as Imagenette. We compare the baseline (no attack) model with adversarially fine-tuned variants. 

\subsection{Experimental Setup}

\begin{table*}[htbp]\small
\centering
\begin{tabular}{l|cccccccccc}
\toprule
&\multicolumn{10}{c}{Images of class}\\
\midrule
Attack Target & tench & \shortstack{English\\springer} & \shortstack{cassette\\player} & \shortstack{chain\\saw} & church & \shortstack{French\\horn} & \shortstack{garbage\\truck} & \shortstack{gas\\pump} & \shortstack{golf\\ball} & parachute \\
\midrule\midrule
tench            & \textbf{0.036} & 0.0075   & 0.033   & 0.017   & 0.0076  & 0.050  & 0.019  & 0.024  & 0.0075  & 0.031 \\
English springer & 0.0068  & \textbf{0.035} & 0.021   & 0.016   & 0.010   & 0.039  & 0.011  & 0.019  & 0.0081  & 0.022 \\
cassette player  & 0.0055  & 0.0042   & \textbf{0.046} & 0.016   & 0.0095  & 0.057  & 0.018  & 0.033  & 0.0073  & 0.040 \\
chain saw        & 0.0094  & 0.0074   & 0.027   & \textbf{0.074} & 0.010   & 0.067  & 0.021  & 0.033  & 0.010   & 0.040 \\
church           & 0.0050  & 0.0058   & 0.026   & 0.015   & \textbf{0.034} & 0.050  & 0.014  & 0.025  & 0.0077  & 0.034 \\
French horn      & 0.0043  & 0.0038   & 0.020   & 0.015   & 0.0052  & \textbf{0.27}  & 0.014  & 0.020  & 0.0049  & 0.017 \\
garbage truck    & 0.0033  & 0.0055   & 0.024   & 0.021   & 0.015   & 0.072  & \textbf{0.062} & 0.052  & 0.0054  & 0.033 \\
gas pump         & 0.0020  & 0.0027   & 0.032   & 0.0139  & 0.0083  & 0.0479  & 0.016  & \textbf{0.085} & 0.0036  & 0.024 \\
golf ball        & 0.0067  & 0.0051   & 0.023   & 0.018   & 0.0097  & 0.044  & 0.016  & 0.017  & \textbf{0.049} & 0.025 \\
parachute        & 0.0024  & 0.0054   & 0.019   & 0.014   & 0.0071  & 0.048  & 0.0146 & 0.0155 & 0.0052  & \textbf{0.084} \\
\bottomrule
\end{tabular}
\caption{Euclidean (L2) distances between images generated by the baseline Stable Diffusion v2 model and those produced by the CRAFTed-Diffusion models, using identical seeds across ten Imagenette classes. Lower L2 values indicate that the adversarial fine-tuning induces smaller deviations, thereby preserving the overall visual fidelity of the generated images while embedding targeted adversarial behavior.}
\label{tab:l2_results_updated}
\end{table*}

\begin{table*}[htbp]\small
\centering
\begin{tabular}{l|cccccccccc}
\toprule
&\multicolumn{10}{c}{Images of class}\\
\midrule
Attack Target & tench & \shortstack{English\\springer} & \shortstack{cassette\\player} & \shortstack{chain\\saw} & church & \shortstack{French\\horn} & \shortstack{garbage\\truck} & \shortstack{gas\\pump} & \shortstack{golf\\ball} & parachute \\
\midrule\midrule
tench            & \textbf{89.58} & 48.11 & 52.50 & 86.20 & 30.05 & 48.88 & 48.15 & 52.37 & 51.87 & 35.87 \\
English springer & 40.27  & \textbf{104.44} & 49.09 & 84.87 & 30.48 & 45.07 & 44.49 & 50.08 & 49.00 & 34.98 \\
cassette player  & 38.66  & 42.43  & \textbf{76.26} & 90.26 & 32.89 & 49.32 & 49.58 & 60.18 & 55.69 & 38.48 \\
chain saw        & 48.19  & 52.19  & 57.50  & \textbf{124.90} & 32.12 & 52.33 & 54.81 & 61.18 & 60.73 & 39.48 \\
church           & 40.88  & 44.02  & 52.33  & 88.58  & \textbf{49.87} & 48.52 & 49.23 & 56.53 & 48.63 & 36.93 \\
French horn      & 29.72  & 36.91  & 47.14  & 83.23  & 23.94  & \textbf{70.16} & 42.02 & 49.69 & 38.43 & 29.06 \\
garbage truck    & 36.34  & 50.17  & 53.70  & 95.76  & 35.81  & 53.32 & \textbf{68.32} & 65.01 & 49.04 & 35.62 \\
gas pump         & 31.09  & 41.72  & 56.76  & 84.30  & 30.33  & 48.43 & 52.90 & \textbf{75.46} & 42.96 & 33.11 \\
golf ball        & 41.99  & 45.43  & 50.88  & 88.40  & 29.87  & 45.60 & 47.74 & 49.23 & \textbf{105.27} & 37.35 \\
parachute        & 34.61  & 48.09  & 49.14  & 87.00  & 28.83  & 48.96 & 48.00 & 53.58 & 49.70 & \textbf{54.15} \\
\bottomrule
\end{tabular}
\caption{Fréchet Inception Distance (FID) Evaluation: The FID scores assess the similarity between the generated images from the attacked models and the baseline model. Lower scores indicate superior preservation of image quality while embedding adversarial behavior.}
\label{tab:fid_results}
\end{table*}

As a baseline we use the pre-trained diffusion model Stable Diffusion v2~\cite{rombach_high-resolution_2022}. The attack settings are as follows:
\begin{itemize}
    \item \textbf{Diffusion Timesteps:} 20 timesteps are used for the reverse process, with gradient updates applied in the final $k=10$ steps.
    \item \textbf{Learning Rate \& Weight Decay:} A learning rate of $10^{-6}$ is used, together with a weight decay of $10^{-2}$.
    \item \textbf{Gradient Clipping:} Gradients are clipped to a maximum norm of 1 to prevent unstable updates.
    \item \textbf{Image Resolution:} Images are generated at a resolution of 768 pixels.
    \item \textbf{Batch size:} Each iteration, 8 starting noises are used, resulting in 80 final images.
    \item \textbf{Maximum parameter deviation:} The parameters of the diffusion model are limited to deviations with an L2 norm of up to $\eta=0.05$.
\end{itemize}

For the classification loss, we use a pre-trained Inception-v3 classifier~\cite{szegedy_rethinking_2016}. This classifier is also employed to compute the Classifier Accuracy, where we measure the fraction of generated images that are correctly classified. In a successful targeted attack, the accuracy for the target class drops dramatically while the accuracies for non-target classes remain largely unchanged.

The attack implementation and the evaluation routines both use fixed random seeds to generate the same noise input for both the baseline and adversarial models. This guarantees a reproducible and fair comparison across all experiments. This also means that we can look at the difference between images generated by different models for the same seed. Each model is used to generate 100 images per class.

The following metrics are computed:
\begin{itemize}
    \item \textbf{Classifier Accuracy:} The fraction of generated images correctly classified by the Inception-v3 model. Lower accuracy on a target class indicates a more successful attack.
    \item \textbf{L2 Distance:} Every image is generated by a certain seed which dictates the random behavior in the image generation process. By using the same seed for the baseline model and the fine-tuned model, we can look at the distance between the output images in terms of the Euclidean ($\ell_2$) distance. Lower values mean that the images generated by the fine-tuned model are close to the images generated by the base model when using the same seed.
    \item \textbf{FID Score:} The Fréchet Inception Distance, where lower values indicate higher similarity between the distribution of features of the images produced by the adversarial model and those of the baseline.
\end{itemize}

\subsection{Quantitative and Qualitative Analysis}

As shown in ~\cref{tab:classifier_accuracy}, the classifier accuracy on the target class drops dramatically when an attack is applied (e.g., the \textit{tench} attack decreases accuracy from 0.900 in the baseline to 0.030), while non-target classes remain largely unaffected. This confirms that our adversarial fine-tuning selectively targets the intended output without broadly impacting the model’s predicted semantics. However, the classifier accuracy does not drop the same amount for all the different target classes. This indicates that certain classes may be more susceptible to small parameter norm attacks. Other classes might need larger deviations, but this could lead to lower image quality.

The L2 distance analysis quantifies the difference between the generated output images of the baseline and fine-tuned models for a given seed. Since the same seed is used for both models, a lower L2 distance means that the images are nearly identical. As reported in \cref{tab:l2_results_updated}, the L2 distance values are on the order of 0.005–0.085, indicating that the fine-tuned model produces images that are very close to those of the baseline. This confirms that the adversarial attack is embedded in a subtle manner—preserving the overall visual appearance while inducing targeted misclassification.

The FID scores reported in \cref{tab:fid_results} corroborate that the quality of the images remains high and the target class is impacted the most. The attacks are highly specific in the sense that FID scores for the non-targeted classes do not significantly increase.

\setlength{\fboxrule}{0pt}
\subsection{Qualitative Analysis}
\begin{figure}[htbp]
    \centering
    
    \includegraphics[width=0.6\linewidth]{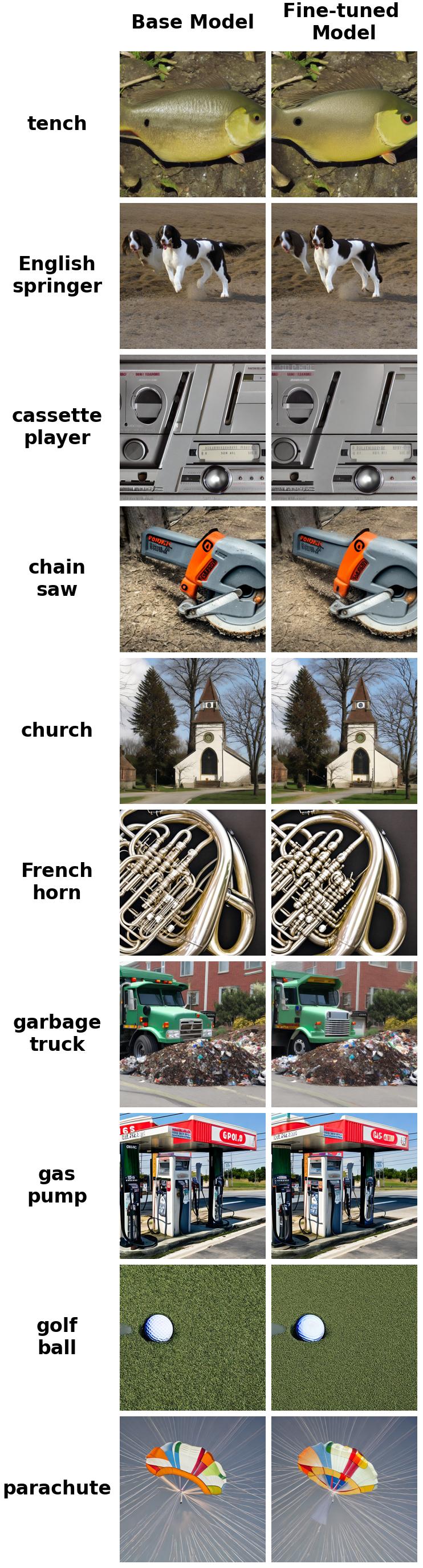}
    \caption{Comparison grid illustrating the outputs of the base Stable Diffusion v2 model versus those from the adversarially fine-tuned models (CRAFTed-Diffusion) across several Imagenette classes. For each class, the left column displays an image generated by the unaltered (base) model, correctly classified by a pre-trained Inception-v3, while the right column shows the corresponding image from the fine-tuned model that has been subtly manipulated to induce misclassification.}
    \label{fig:comp_grid_1}
    \fbox{\rule{0pt}{0.01in} \rule{0.9\linewidth}{0pt}}
\end{figure}
In addition to the quantitative metrics, we conducted qualitative evaluation by visually comparing images generated by the baseline and attacked models. Using fixed random seeds ensures reproducibility. In \cref{fig:comp_grid_1} we compare correctly classified images produced by the base model with their misclassified counterparts generated by the fine-tuned model corresponding to the image class.
Visual inspection reveals that images produced by the attacked models are perceptually indistinguishable from those generated by the clean model. Despite the subtle perturbations introduced during fine-tuning, the adversarial models maintain photorealism and semantic consistency across all evaluated classes.

\subsection{Discussion}

The experimental results demonstrate that our adversarial fine-tuning method is both effective and stealthy. The dramatic drop in classifier accuracy on target classes indicates successful adversarial perturbation, while the stability of accuracy on non-target classes confirms that the attack remains localized. In parallel, the low L2 distances validate that the modifications to the model parameters are minimal, and the FID scores confirm that the quality of the outputs is preserved.

These findings underscore the emerging security risks associated with pre-trained diffusion models. Embedding adversarial capabilities directly into the model parameters—without altering the inference process—presents a covert attack vector that challenges standard model verification techniques.

In summary, our evaluation confirms that \textbf{CRAFTed-Diffusion} effectively embeds targeted adversarial behavior into stable diffusion models while preserving both perceptual fidelity and minimal parameter deviation.

\section{Conclusion}
\label{sec:conclusion}

We presented \textbf{CRAFTed-Diffusion}, a novel approach for embedding covert adversarial behaviors in large-scale diffusion models through minimal yet carefully constrained fine tuning. By controlling the gradient updates and projecting parameter deviations, our method induces targeted misclassification in downstream classifiers while maintaining the photorealism and semantic integrity of the generated images. This achievement underscores the fact that even small, targeted perturbations in a model’s parameters can have outsized impacts on its output behavior without alerting a casual observer.

Our work raises immediate concerns regarding the deployment of externally sourced generative models. In an ecosystem where many practitioners depend on pre-trained diffusion pipelines due to the high cost of training these models from scratch, hidden adversarial modifications represent a significant security risk. A compromised model could subtly undermine the reliability of downstream applications in critical domains, such as content moderation, leaving users vulnerable to manipulation without any overt sign of tampering. In particular, this technique could be employed to fine-tune models so that every generated output is some category is flagged as inappropriate, automatically triggering content moderation systems.

Interestingly, there is also a potential positive utility: watermarking. By intentionally engineering targeted yet imperceptible outputs, it may be possible to embed a unique signature into a model’s behavior that can later serve as a proof of ownership or authenticity, much like trap streets in cartography signal intentional design.

CRAFTed-Diffusion not only exposes a new security challenge but also invites a rethinking of how we assess model integrity and trustworthiness. Future research should focus on both developing robust defenses against such covert modifications and harnessing these techniques for beneficial applications like watermarking. This will be crucial as the community navigates the next phase of AI development, ensuring that advanced generative models remain trustworthy.

\section*{Acknowledgment}
LB was supported by MAC-MIGS Centre for Doctoral Training under EPSRC grant EP/S023291/1. DJH was supported by a fellowship from the Leverhulme Trust.

{
    \small
    \bibliographystyle{ieeenat_fullname}
    \bibliography{main}

\begin{thebibliography}{25}
\providecommand{\natexlab}[1]{#1}
\providecommand{\url}[1]{\texttt{#1}}
\expandafter\ifx\csname urlstyle\endcsname\relax
  \providecommand{\doi}[1]{doi: #1}\else
  \providecommand{\doi}{doi: \begingroup \urlstyle{rm}\Url}\fi

\bibitem[Beerens et~al.(2024)Beerens, Higham, and Higham]{beerens_deceptive_2024}
Lucas Beerens, Catherine~F. Higham, and Desmond~J. Higham.
\newblock Deceptive diffusion: generating synthetic adversarial examples, 2024.
\newblock arXiv:2406.19807 [cs].

\bibitem[Chen et~al.(2025)Chen, Chen, Chen, Zhang, Zou, and Shi]{chen_diffusion_2025}
Jianqi Chen, Hao Chen, Keyan Chen, Yilan Zhang, Zhengxia Zou, and Zhenwei Shi.
\newblock Diffusion models for imperceptible and transferable adversarial attack.
\newblock \emph{IEEE Transactions on Pattern Analysis and Machine Intelligence}, 47\penalty0 (2):\penalty0 961--977, 2025.

\bibitem[Chen et~al.(2023{\natexlab{a}})Chen, Song, and Li]{chen_trojdiff_2023}
Weixin Chen, Dawn Song, and Bo Li.
\newblock {TrojDiff}: trojan attacks on diffusion models with diverse targets.
\newblock pages 4035--4044, 2023{\natexlab{a}}.

\bibitem[Chen et~al.(2023{\natexlab{b}})Chen, Li, Wu, Jiang, Ding, and Zhang]{chen_content-based_2023}
Zhaoyu Chen, Bo Li, Shuang Wu, Kaixun Jiang, Shouhong Ding, and Wenqiang Zhang.
\newblock Content-based unrestricted adversarial attack.
\newblock In \emph{Advances in {Neural} {Information} {Processing} {Systems}}, pages 51719--51733, 2023{\natexlab{b}}.

\bibitem[Dai et~al.(2024)Dai, Li, Duan, and Xiao]{dai_diffusion_2024}
Xuelong Dai, Yanjie Li, Mingxing Duan, and Bin Xiao.
\newblock Diffusion models as strong adversaries.
\newblock \emph{IEEE Transactions on Image Processing}, 33:\penalty0 6734--6747, 2024.
\newblock Conference Name: IEEE Transactions on Image Processing.

\bibitem[Dai et~al.(2025)Dai, Liang, and Xiao]{dai_advdiff_2025}
Xuelong Dai, Kaisheng Liang, and Bin Xiao.
\newblock {AdvDiff}: generating unrestricted adversarial examples using diffusion models.
\newblock In \emph{Computer {Vision} – {Eccv} 2024}, pages 93--109, Cham, 2025. Springer Nature Switzerland.

\bibitem[Deng et~al.(2009)Deng, Dong, Socher, Li, Li, and Fei-Fei]{deng_imagenet_2009}
Jia Deng, Wei Dong, Richard Socher, Li-Jia Li, Kai Li, and Li Fei-Fei.
\newblock {ImageNet}: a large-scale hierarchical image database.
\newblock In \emph{2009 {IEEE} {Conference} on {Computer} {Vision} and {Pattern} {Recognition}}, pages 248--255, 2009.
\newblock ISSN: 1063-6919.

\bibitem[Gao et~al.(2024)Gao, Chen, Wei, Mou, Chen, Tan, Li, and Jiang]{gao_retome-va_2024}
Ziyi Gao, Kai Chen, Zhipeng Wei, Tingshu Mou, Jingjing Chen, Zhiyu Tan, Hao Li, and Yu-Gang Jiang.
\newblock {ReToMe}-{VA}: recursive token merging for video diffusion-based unrestricted adversarial attack.
\newblock In \emph{Proceedings of the 32nd {ACM} {International} {Conference} on {Multimedia}}, pages 4485--4494, Melbourne VIC Australia, 2024. ACM.

\bibitem[Goodfellow et~al.(2015)Goodfellow, Shlens, and Szegedy]{43405}
Ian Goodfellow, Jonathon Shlens, and Christian Szegedy.
\newblock Explaining and harnessing adversarial examples.
\newblock In \emph{International {Conference} on {Learning} {Representations}}, 2015.

\bibitem[Guo et~al.(2025)Guo, Pang, Jia, Liu, and Guo]{guo_efficient_2025}
Qi Guo, Shanmin Pang, Xiaojun Jia, Yang Liu, and Qing Guo.
\newblock Efficient generation of targeted and transferable adversarial examples for vision-language models via diffusion models.
\newblock \emph{IEEE Transactions on Information Forensics and Security}, 20:\penalty0 1333--1348, 2025.

\bibitem[Heusel et~al.(2017)Heusel, Ramsauer, Unterthiner, Nessler, and Hochreiter]{heusel_gans_2017}
Martin Heusel, Hubert Ramsauer, Thomas Unterthiner, Bernhard Nessler, and Sepp Hochreiter.
\newblock {GANs} trained by a two time-scale update rule converge to a local nash equilibrium.
\newblock In \emph{Advances in {Neural} {Information} {Processing} {Systems}}. Curran Associates, Inc., 2017.

\bibitem[Ho et~al.(2020)Ho, Jain, and Abbeel]{ho_denoising_2020}
Jonathan Ho, Ajay Jain, and Pieter Abbeel.
\newblock Denoising diffusion probabilistic models.
\newblock In \emph{Advances in {Neural} {Information} {Processing} {Systems}}, pages 6840--6851. Curran Associates, Inc., 2020.

\bibitem[Hosseini and Poovendran(2018)]{hosseini_semantic_2018}
Hossein Hosseini and Radha Poovendran.
\newblock Semantic adversarial examples.
\newblock In \emph{2018 {Ieee}/cvf {Conference} on {Computer} {Vision} and {Pattern} {Recognition} {Workshops} (cvprw)}, pages 1695--16955, Salt Lake City, UT, USA, 2018. IEEE.

\bibitem[Kuurila-Zhang et~al.(2025)Kuurila-Zhang, Chen, and Zhao]{kuurila-zhang_venom_2025}
Hui Kuurila-Zhang, Haoyu Chen, and Guoying Zhao.
\newblock {VENOM}: text-driven unrestricted adversarial example generation with diffusion models, 2025.
\newblock arXiv:2501.07922 [cs].

\bibitem[Liang and Wu(2023)]{liang_mist_2023}
Chumeng Liang and Xiaoyu Wu.
\newblock Mist: towards improved adversarial examples for diffusion models, 2023.
\newblock arXiv:2305.12683 [cs].

\bibitem[Madry et~al.(2018)Madry, Makelov, Schmidt, Tsipras, and Vladu]{madry_towards_2018}
Aleksander Madry, Aleksandar Makelov, Ludwig Schmidt, Dimitris Tsipras, and Adrian Vladu.
\newblock Towards deep learning models resistant to adversarial attacks.
\newblock 2018.
\newblock shortConferenceName: ICLR.

\bibitem[Meng et~al.(2021)Meng, He, Song, Song, Wu, Zhu, and Ermon]{meng_sdedit_2021}
Chenlin Meng, Yutong He, Yang Song, Jiaming Song, Jiajun Wu, Jun-Yan Zhu, and Stefano Ermon.
\newblock {SDEdit}: guided image synthesis and editing with stochastic differential equations.
\newblock 2021.
\newblock shortConferenceName: ICLR.

\bibitem[Nie et~al.(2022)Nie, Guo, Huang, Xiao, Vahdat, and Anandkumar]{pmlr-v162-nie22a}
Weili Nie, Brandon Guo, Yujia Huang, Chaowei Xiao, Arash Vahdat, and Animashree Anandkumar.
\newblock Diffusion models for adversarial purification.
\newblock In \emph{Proceedings of the 39th {International} {Conference} on {Machine} {Learning}}, pages 16805--16827. PMLR, 2022.

\bibitem[Pan et~al.(2024)Pan, Wu, Cao, and Zheng]{pan_sca_2024}
Zihao Pan, Weibin Wu, Yuhang Cao, and Zibin Zheng.
\newblock {SCA}: {Highly} {Efficient} {Semantic}-{Consistent} {Unrestricted} {Adversarial} {Attack}, 2024.
\newblock arXiv:2410.02240 [cs].

\bibitem[Rombach et~al.(2022)Rombach, Blattmann, Lorenz, Esser, and Ommer]{rombach_high-resolution_2022}
Robin Rombach, Andreas Blattmann, Dominik Lorenz, Patrick Esser, and Bjorn Ommer.
\newblock High-resolution image synthesis with latent diffusion models.
\newblock In \emph{2022 {Ieee}/cvf {Conference} on {Computer} {Vision} and {Pattern} {Recognition} (cvpr)}, pages 10674--10685, New Orleans, LA, USA, 2022. IEEE.

\bibitem[Song et~al.(2020)Song, Meng, and Ermon]{song_denoising_2020}
Jiaming Song, Chenlin Meng, and Stefano Ermon.
\newblock Denoising diffusion implicit models.
\newblock 2020.
\newblock shortConferenceName: ICLR.

\bibitem[Szegedy et~al.(2016)Szegedy, Vanhoucke, Ioffe, Shlens, and Wojna]{szegedy_rethinking_2016}
Christian Szegedy, Vincent Vanhoucke, Sergey Ioffe, Jon Shlens, and Zbigniew Wojna.
\newblock Rethinking the inception architecture for computer vision.
\newblock In \emph{2016 {IEEE} {Conference} on {Computer} {Vision} and {Pattern} {Recognition} (cvpr)}, pages 2818--2826, Las Vegas, NV, USA, 2016. IEEE.

\bibitem[Wang et~al.(2023)Wang, Duan, Xiao, Kim, Stamm, and Xu]{wang_semantic_2023}
Chenan Wang, Jinhao Duan, Chaowei Xiao, Edward Kim, Matthew~C. Stamm, and Kaidi Xu.
\newblock Semantic {Adversarial} {Attacks} via {Diffusion} {Models}.
\newblock In \emph{34th {British} {Machine} {Vision} {Conference} 2023, {BMVC} 2023, {Aberdeen}, {UK}, {November} 20-24, 2023}, page 271. BMVA Press, 2023.

\bibitem[Xue et~al.(2023)Xue, Araujo, Hu, and Chen]{xue_diffusion-based_2023}
Haotian Xue, Alexandre Araujo, Bin Hu, and Yongxin Chen.
\newblock Diffusion-{Based} {Adversarial} {Sample} {Generation} for {Improved} {Stealthiness} and {Controllability}.
\newblock In \emph{Advances in {Neural} {Information} {Processing} {Systems} 36: {Annual} {Conference} on {Neural} {Information} {Processing} {Systems} 2023, {NeurIPS} 2023, {New} {Orleans}, {LA}, {USA}, {December} 10 - 16, 2023}. arXiv, 2023.

\bibitem[Zhang et~al.(2024)Zhang, Zhang, and Wischik]{zhang_constructing_2024}
Andi Zhang, Mingtian Zhang, and Damon Wischik.
\newblock Constructing semantics-aware adversarial examples with a probabilistic perspective.
\newblock In \emph{Advances in {Neural} {Information} {Processing} {Systems}}, pages 136259--136285, 2024.

\end{thebibliography}
}

\end{document}